\documentclass[letterpaper, 10 pt, conference]{ieeeconf}  

\IEEEoverridecommandlockouts                              

\usepackage{cite}
\usepackage{amsmath,amssymb,amsfonts}
\usepackage[boxruled, noend]{algorithm2e}
\usepackage{graphicx}
\usepackage{float}
\usepackage{textcomp}
\usepackage{xcolor}
\def\BibTeX{{\rm B\kern-.05em{\sc i\kern-.025em b}\kern-.08em
    T\kern-.1667em\lower.7ex\hbox{E}\kern-.125emX}}

\graphicspath{{Figures/}}

\usepackage[font={small}]{caption}
\newcommand{\reffig}[1]{Fig.~\ref{#1}}
\newcommand{\refeq}[1]{\eqref{#1}}
\newcommand{\reftab}[1]{Tab.~\ref{#1}}
\newcommand{\myss}[1]{\mathrm{#1}}
\newcommand{\mat}[1]{\mathbf{#1}}
\newcommand{\LIDAR}{LIDAR}

\title{\LARGE \bf
Deep-Learning Assisted High-Resolution Binocular Stereo Depth Reconstruction
}

\author{Yaoyu Hu$^{1}$, Weikun Zhen$^{2}$, and Sebastian Scherer$^{3}$
\thanks{*This work is supported by the Shimizu Corporation}
\thanks{$^{1,3}$Yaoyu Hu and Sebastian Scherer are with the Robotics Institute, Carnegie Mellon University, Pittsburgh, PA 15213, USA
        {\tt\small yaoyuh@andrew.cmu.edu; basti@andrew.cmu.edu}}%
\thanks{$^{2}$Weikun Zhen is with the Department of Mechanical Engineering, Carnegie Mellon University, Pittsburgh, PA 15213, USA
        {\tt\small weikunz@andrew.cmu.edu}}%
}

\begin{document}

\maketitle
\thispagestyle{empty}
\pagestyle{empty}

\begin{abstract}
This work presents dense stereo reconstruction using high-resolution images for infrastructure inspections. The state-of-the-art stereo reconstruction methods, both learning and non-learning ones, consume too much computational resource on high-resolution data. Recent learning-based methods achieve top ranks on most benchmarks. However, they suffer from the generalization issue due to lack of task-specific training data. We propose to use a less resource demanding non-learning method, guided by a learning-based model, to handle high-resolution images and achieve accurate stereo reconstruction. The deep-learning model produces an initial disparity prediction with uncertainty for each pixel of the down-sampled stereo image pair. The uncertainty serves as a self-measurement of its generalization ability and the per-pixel searching range around the initially predicted disparity. The downstream process performs a modified version of the Semi-Global Block Matching method with the up-sampled per-pixel searching range. The proposed deep-learning assisted method is evaluated on the Middlebury dataset and high-resolution stereo images collected by our customized binocular stereo camera. The combination of learning and non-learning methods achieves better performance on 12 out of 15 cases of the Middlebury dataset. In our infrastructure inspection experiments, the average 3D reconstruction error is less than 0.004m.

\end{abstract}

\section{Introduction}
\begin{figure}[htp]
    \centering
    \includegraphics[width=3in]{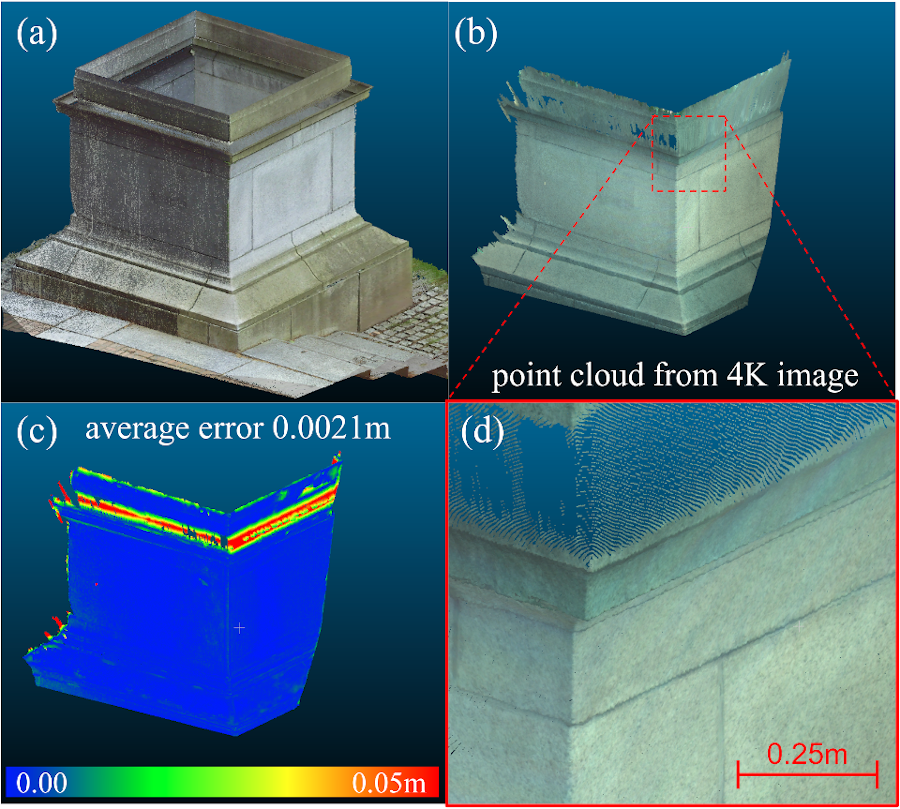}
    \caption{Reconstructed point cloud from 4K resolution stereo images. (a) Point cloud from a survey scanner, used as true data for comparison. (b) Densely-reconstructed point cloud by our deep-learning assisted method. (c) Reconstruction error of (b) compared with (a). (d) Zoomed view of (b). }
    \label{fig:cover_figure}
\end{figure}

There is a widespread integration of UAV (Unmanned Aerial Vehicle) technology in the infrastructure inspection, which requires dense 3D reconstruction of facilities such as bridges and power grids. 
Binocular stereo camera is widely used for dense depth reconstruction due to its simplicity and low-cost hardware. However, stereo matching for high-resolution images is still a challenging task because of the huge amount of computation brought by the large disparity searching range and pixel number. In addition, image-based reconstructions suffer from lack of texture, slanted surfaces \cite{bleyer2011patchmatch} and inadequate lighting, leading to reconstruction failure and sparse disparity predictions in these difficult image regions. 
Our goal is dense and accurate depth reconstruction for high-resolution images as shown in \reffig{fig:cover_figure}. 

High-resolution stereo images have a large number of pixels and large searching ranges for potential disparities. Recently, deep-learning methods tend to out-perform non-learning ones. However, the image size is limited by both the available training data and computing hardware. Image sizes of common binocular stereo datasets are under 1 megapixel, e.g, KITTI 2015 \cite{Menze2015CVPR}, Scene Flow \cite{MIFDB16} and NYU \cite{silberman2012indoor}. 
Techniques such as encoder-decoder with skip connections \cite{he2016deep} and spacial polling \cite{liu2015parsenet, zhao2017pyramid} could reduce memory requirements. However, most of the methods also adopt the cost volume concept \cite{kendall2017end} which consumes substantial amount of GPU memory. Recent works reduce memory even further, e.g. \cite{Yao_2019_CVPR}, but still not enough to fit a pair of our 4K images. 

Non-learning methods such as the Semi-Global Matching (SGM) consumes over 50GB of CPU memory on 4K images (12 megapixels) with a disparity range of 1000 (measured from the SGM part of SPS-Stereo \cite{yamaguchi2014efficient}). The SGBM (Semi-Global Block Matching, which runs a simplified version of SGM by default) method implemented in the OpenCV \cite{OpenCV} package could handle our 4K image pairs directly with restricted computing resources. However, the model parameters are case dependent and SGBM might fail to predict disparities in many image regions as shown later in \reffig{fig:exp_comparison_row}. In these failure regions, searching for a stereo match within a large disparity range is difficult because there may be multiple disparities with similar matching cost. Intuitively, the search may be easier if the range is narrowed. 

We also observed that a deep-learning model could estimate its uncertainty of the disparity prediction\cite{kendall2019geometry}. The uncertainty is a good hint of a possible disparity range which narrows the disparity searching range. Our work follows the above observations. The key contributions are:

\begin{itemize}
    \item We propose a hybrid approach that uses a learning-based model to guide a non-learning method in order to achieve high efficiency and accuracy in high-resolution stereo reconstruction tasks. 
    \item We train a deep-learning model to produce both disparity and uncertainty. The uncertainty is further utilized as the per-pixel searching range by a non-learning method.
    \item We show in the experiments that the combination of learning and non-learning methods could accurately process high-resolution stereo images. 
\end{itemize}


\section{Related work}
The majority of the recent work related to depth estimation use deep-learning models. Some are targeting high-resolution images, e.g. Pillai et al. \cite{pillai2018superdepth} apply subpixel-convolutional layers to an encoder-decoder architecture to deal with large images. Wang et al. \cite{wang2019anytime} propose to initially predict depth from down-sampled images. Then the depth is incrementally up-scaled and corrected by a deep-learning model. Wofk et al. \cite{wofk2019fastdepth} perform network pruning to reduce the resource requirement. These models could not handle the 4K image size of our stereo camera. Besides, high-resolution training data are not available. Considering the above limitations, we train and test our deep-learning model on low-resolution stereo images. 

Our approach needs a deep-learning model to estimate uncertainty of its disparity prediction. Our inspiration is derived from the work of Gal \cite{gal2016uncertainty} and Kendal \cite{kendall2019geometry} on uncertainties of deep-learning methods. According to them and \cite{der2009aleatory}, a model could learn to estimate the aleatoric uncertainty which partially depends on the individual input data. 
During inference, the trained model predicts the possible error it might make on the current input. Our work embraces this method and makes a deep-learning model predict per-pixel disparity uncertainty. Then the uncertainty is directly utilized to compute the possible disparity range.

In our proposed approach, we guide an SGM algorithm and achieve better performance by directly providing depth estimation. Similar methods are utilized to do depth completion or sensor fusion for stereo vision. Most of the fused depth information comes from direct sparse measurement, such as \LIDAR \cite{premebida2016high, courtois2017fusion, uhrig2017sparsity, mal2018sparse} and ToF (Time-of-Flight) sensors \cite{nair2013survey}. In our work, the depth information comes from a deep-learning model on a per-pixel basis. Like Fischer et al.\cite{fischer2011combination}, we are going to modify the cost aggregation process \cite{scharstein2002taxonomy} of SGM with the implementation inspired by the work of Shivakumar et al.\cite{shivakumar2019real}.

\section{Technical Approach}

\begin{figure}[htb]
    \centering
    \includegraphics[width=3.2in]{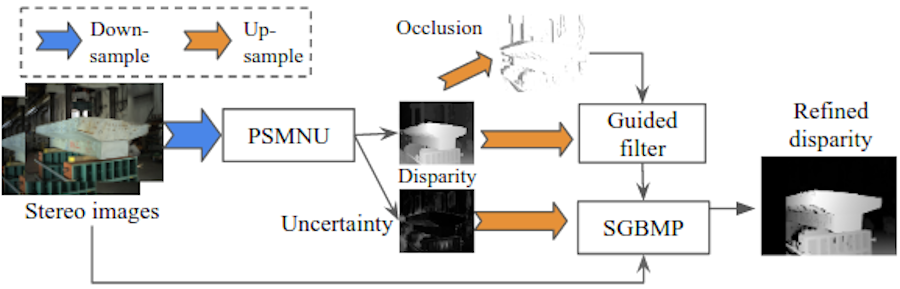}
    \caption{Processing pipeline of the proposed approach. PSMNU: our deep-learning model. SGBMP: deep-learning assisted non-learning model. PSMNU predicts the disparity and uncertainty on the down-sampled stereo images. The disparity is up-sampled to the original size. Occlusions are derived from the disparity. Disparity and occlusion proposals go through a special-purpose filter. SGBMP uses the filtered data and the up-sampled uncertainty to predict the refined disparity with the original stereo images.}
    \label{fig:flow_chart}
\end{figure}

\reffig{fig:flow_chart} shows the processing pipeline of the proposed approach. A deep-learning model (referred to as PSMNU) predicts both disparity and uncertainty from down-sampled images. The initial disparity is up-sampled to the original resolution. Then an occlusion proposal is derived from the disparity. The disparity and the occlusion proposal are processed by a guided filter. The uncertainty estimation is used to determine the per-pixel searching range (PPSR) which is also up-sampled. Finally, a modified SGBM algorithm (referred to as SGBMP) takes in the filtered disparity, occlusion proposal, and PPSR to weight the aggregated matching cost. With this pipeline, we recover accurate dense disparity predictions for 4K stereo images.

\subsection{Deep-learning model with uncertainty}

We build PSMNU based on the PSMNet \cite{chang2018pyramid} which has promising performance. Following \cite{kendall2019geometry}, we modify the PSMNet to predict aleatoric uncertainties. Let $\mathbf{f}$ represent our deep-learning model as a function $\mathbf{f}(\mat{x})$ which maps stereo images $\mat{x}$ to disparity of the left image. This mapping could be considered to be a random process and is assumed to follow a per-pixel Gaussian distribution expressed in \refeq{eq:gaussian}.
\begin{equation} \label{eq:gaussian}
    P\left( y_{p}=y^{\myss{t}}_{p} | f_{p}\left ( \mat{x} \right ) \right) = \mathcal{N} \left( f_{p} ( \mat{x}), \sigma^{2}_{p}\left ( \mat{x} \right) \right )
\end{equation}

\noindent where $y_{p}$ is our disparity prediction for pixel $p$ and $y^{\myss{t}}_{p}$ is the true disparity. $f_p$ denotes a mapping which produces a disparity at pixel $p$. The probability density of our model predicting a $y_{p}$ equal to $y^{\myss{t}}_{p}$ upon seeing $ f_{p}(\mat{x})$ is represented as $P(\cdot)$. $\sigma_p$ is the standard deviation of the Gaussian distribution at pixel $p$. $\sigma_{p}$ also represents the uncertainty. We change the last regression layer of PSMNet to make $\mathbf{f}$ output two channels. One channel is for $y_{p}$ and the other for $\sigma_p$. We refer to our model as PSMNU (PSMNet with Uncertainty) in this work. To stabilize the computation and avoid division by zero, $\delta_p=\log\left( \sigma^2_p \right)$ is produced in practice \cite{kendall2019geometry}. By using the loss function defined in \refeq{eq:loss_function}, PSMNU needs no ground truth for $\delta_p$. 
\begin{equation} \label{eq:loss_function}
    \mathcal{L} = \frac{1}{2N_{\mathbf{p}}} \sum_{p \in \mathbf{p}} E_{p} e^{-\delta_p} + \frac{1}{2N_{\mathbf{p}}} \sum_{\mathbf{p}} \delta_p
\end{equation}

\noindent with $N_{\mat{p}}$ being the number of pixels and $E_p$ defined as
\begin{equation} \label{eq:E_p}
    E_{p} = \left \| y^{\myss{t}}_{p} - f_{p} \left ( \mathbf{x} \right ) \right \|_{2}^2
\end{equation}

In general, a lower $\delta_p$ means more confident. When $\mathbf{f}$ is not confident on $y_p$ then $E_p$ tends to be large. To lower the loss, the model has to predict a large $\delta_p$ to attenuate $E_p$ but regularized by the last term of \refeq{eq:loss_function} which punishes the model from predicting large $\delta_p$ values. In contrast, if a small $E_p$ is predicted, the model is allowed to give a small $\delta_p$ to lower the regularization term and also the loss function. $\delta_p$ (and $\sigma_p$) behaves consistently with its role in the Gaussian distribution defined in \refeq{eq:gaussian}. A large $\delta_p$ leads to uniform and lower probability of $y_{p}$ being equal to $y^{\myss{t}}_{p}$, a small $\delta_p$ indicates that $y_p$ is close to $y^{\myss{t}}_{p}$.  



\begin{figure}[htbp]
    \centering
    \includegraphics[width=3in]{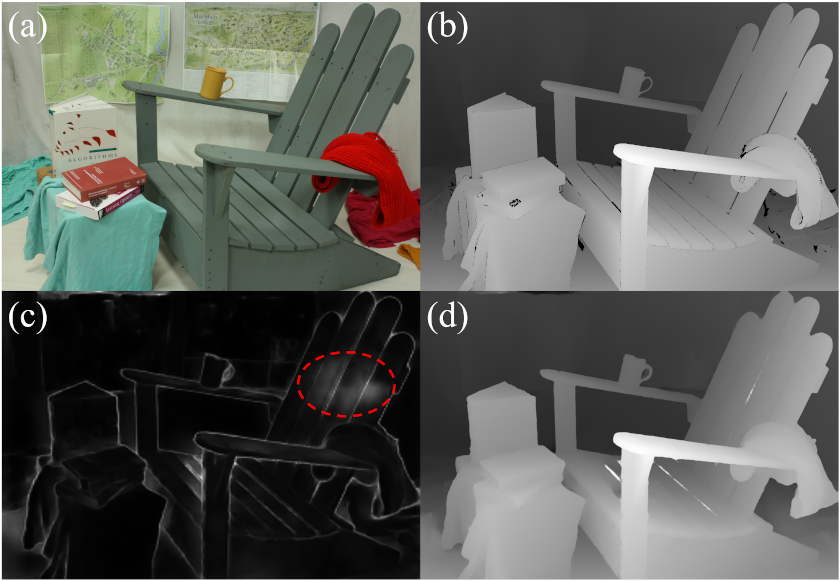}
    \caption{PSMNU output for the Adirondack case of Middlebury dataset. (a) Left image. (b) True disparity. (c) $\sigma$ map, scaled to 0-255 for visualization. (d) Predicted disparity. The red circle in (c) marks a region with high noise level.}
    \label{fig:app_PSMNU_Adirondack}
\end{figure}

We train PSMNU on the Scene Flow dataset (FlyingThings3D, Monkaa) \cite{MIFDB16} with full resolution images and a disparity range of 256. Later, the 4K images will be down-sampled to 1/4 width, and then fed to PSMNU. Therefore, the up-sampled disparity prediction covers 1024 pixels, which is enough for our tests. PSMNU is trained with a mini-batch of 4 on 4 NVIDIA TITAN X GPUs for 5 epochs. A prediction result from PSMNU on Middlebury Stereo Evaluation V3 \cite{scharstein2014high} is shown in \reffig{fig:app_PSMNU_Adirondack} (first down-sampled to 768$\times$1024, result is up-sampled back to full resolution). The prediction has an average error of 1.18 pixels (also shown in \reftab{tab:comparison_middlebury}) compared with the true disparity. The $\sigma$ map shows that high uncertainty exists at most of the object edges where disparities become discontinuous and occlusions happen.  

\subsection{SGBM with per-pixel searching range (PPSR)}

We use the $y_p$ and $\sigma_p$ from PSMNU to determine a disparity range for each pixel. This PPSR is represented as $y_p \pm\lambda^{\myss{b}} \sigma_{p}$ and we set $\lambda^{\myss{b}}=3$ for all our experiments. We implement our method by extending the SGBM method \cite{OpenCVSGBM} of OpenCV and name it SGBMP. Note that $y_p$ and $\sigma_p$ are scaled before the up-sampling to ensure the consistency between image scales. Once obtained the PPSR, we focus on modifying the cost aggregation part of the SGBM method. For each pixel $p$, the SGM aggregation cost of the $i$-th candidate disparity $d_{i,p}$ inside the PPSR is expressed by
\refeq{eq:ssp}.
\begin{equation} \label{eq:ssp}
    S^{\myss{s}}_{p} \left ( d_{i, p} \right ) = S_{p} \lambda^{\myss{s}} \exp \left ( -\lambda^{\myss{d}} \sigma_{p} \left | d_{i, p} - y_{p} \right | \right )
\end{equation}

\noindent where $S_p$ is the aggregated matching cost defined in (14) of \cite{hirschmuller2005accurate},  $S^{\myss{s}}_{p}$ is the weighted $S_p$, $\lambda^{\myss{s}}$ and $\lambda^{\myss{d}}$ are constant parameters. Equation \eqref{eq:ssp} imposes a prior on $S_p$ to favor $y_p$ predicted by PSMNU. However, SGBMP trusts $y_p$ by a discount according to the $\sigma_p$ value. It is further controlled by $\lambda^{\myss{d}}$ and the disparity distance between $d_{i, p}$ and $y_p$. $\lambda^{\myss{s}}$ is a factor which adjusts the global weighting of $S_p$. Typically, $0<\lambda^{\myss{s}}<1$. We empirically decide $\lambda^{\myss{s}} = \lambda^{\myss{d}}=0.1$ in our experiments. In tests with our 4K images, $S_p$ gets saturated easily in some regions due to the large size of the image. SGBM already takes care of this issue following (13) of \cite{hirschmuller2005accurate}. We further scale down the stereo matching cost by a factor of 3 before the cost aggregation.

SGBM applies uniqueness ratio check (or peak ratio check similar to \cite{hu2012quantitative}, referred to as UR check later) and occlusion check to its disparity predictions. UR check is controlled by the \emph{uniquenessRatio} parameter of SGBM. SGBMP disables the UR check. This keeps disparity predictions from being eliminated by the case dependent \emph{uniquenessRatio} in difficult regions such as the areas with low and repetitive texture. SGBMP also does the occlusion check similar to SGBM. However, when $\mat{y}$ from PSMNU has a significant error (such as the region marked by the red circle in \reffig{fig:app_PSMNU_Adirondack}~(c)), the error may survive the occlusion check, leading to occlusions of some other pixels. A special guided filter is developed to process the occlusion proposal derived from PSMNU's initial disparity prediction. The occlusion proposals are represented as a logical mask, $\mat{M}$, with the same size as the left image. Let $\mat{W}(\cdot, j, k)$ be a window which has pixel coordinate $(j, k)$ as the center. Then we run the guided filter described in Algorithm~\ref{alg:guided_filter} in its horizontal version. We run the vertical version filter on the result of the horizontal filtering. The window consists of 3 pixels in a row for the horizontal version and a 3-pixel column for the vertical one. The initial $\mat{y}$ from PSMNU gets updated after the guided filtering. This revised $\mat{y}$ is then used to generate the PPSR for SGBMP. 

\begin{algorithm}[htpb]
    \DontPrintSemicolon
    \SetKwInput{Input}{input}
    \Input{$\mat{y}$, $\mat{M}$}
    Pixel indices use zero-based numbering\;
    \For{$j$-th row of $\mat{y}$ and $\mat{M}$}{
        
        \For{ $k$-th column of $\mat{y}$ and $\mat{M}$, from $k$=1 to $k$=image width - 2 }{
            \If{$\mat{M}$(j, k) is not median of $\mat{W}(\mat{M}, j, k)$}{
                $\mat{y}(j,k) = 0.5\mat{y}(j,k-1) + 0.5\mat{y}(j,k+1)$\;
                $\mat{M}(j,k) = $ \emph{the median of} $\mat{W}(\mat{M}, j, k)$
                }
        }
        
    }

    \caption{Guided filter (horizontal)}
    \label{alg:guided_filter}
\end{algorithm}


\section{Experiments}

\subsection{Comparison with true disparity}

\begin{table*}[ht!]
\caption{Comparison on the Middlebury dataset.}
\label{tab:comparison_middlebury}
\centering
\begin{tabular}{c|ccccc|cccc|cccc}
   \hline
   Case & \multicolumn{5}{c|}{Adirondack} & \multicolumn{4}{c|}{PlaytableP} & \multicolumn{4}{c}{Jadeplant} \\
   \hline
    Metric & bad1.0 & invalid & avgErr & stdErr & time(s) & bad1.0 & invalid & avgErr & stdErr & bad1.0 & invalid & avgErr & stdErr \\
    \hline
   PSMNU  & 20.92 & (7.52) & 1.18 & 4.81 & 6.98    & 38.00 & (6.53) & 7.83 & 22.77                 & 35.09 & (19.35) & 18.29 & 54.90  \\
   SGBM   & \bf{21.59} & 31.60 & 9.54 & 30.54 & 4.89     & \bf{17.46} & 20.79 & 2.15 & 7.30              & \bf{13.90} & 34.26 & \bf{7.54} & 37.45 \\
   SGBMUR & 32.40 & 18.85 & 14.17 & 37.71 & 4.93         & 25.57 & 10.75 & 2.78 & 8.31                   & 19.61 & \bf{27.29} & 11.62 & 46.53 \\
   SGBMP  & 24.82 & \bf{8.47} & \bf{1.10} & \bf{3.61} & 7.19       & 25.01 & \bf{8.65} & \bf{1.68} & \bf{5.19}     & 26.11 & 28.23 & 7.84 & \bf{35.74} \\
   \hline
\end{tabular}
\end{table*}

\begin{figure*}[h]
    \centering
    \includegraphics[width=6.4in]{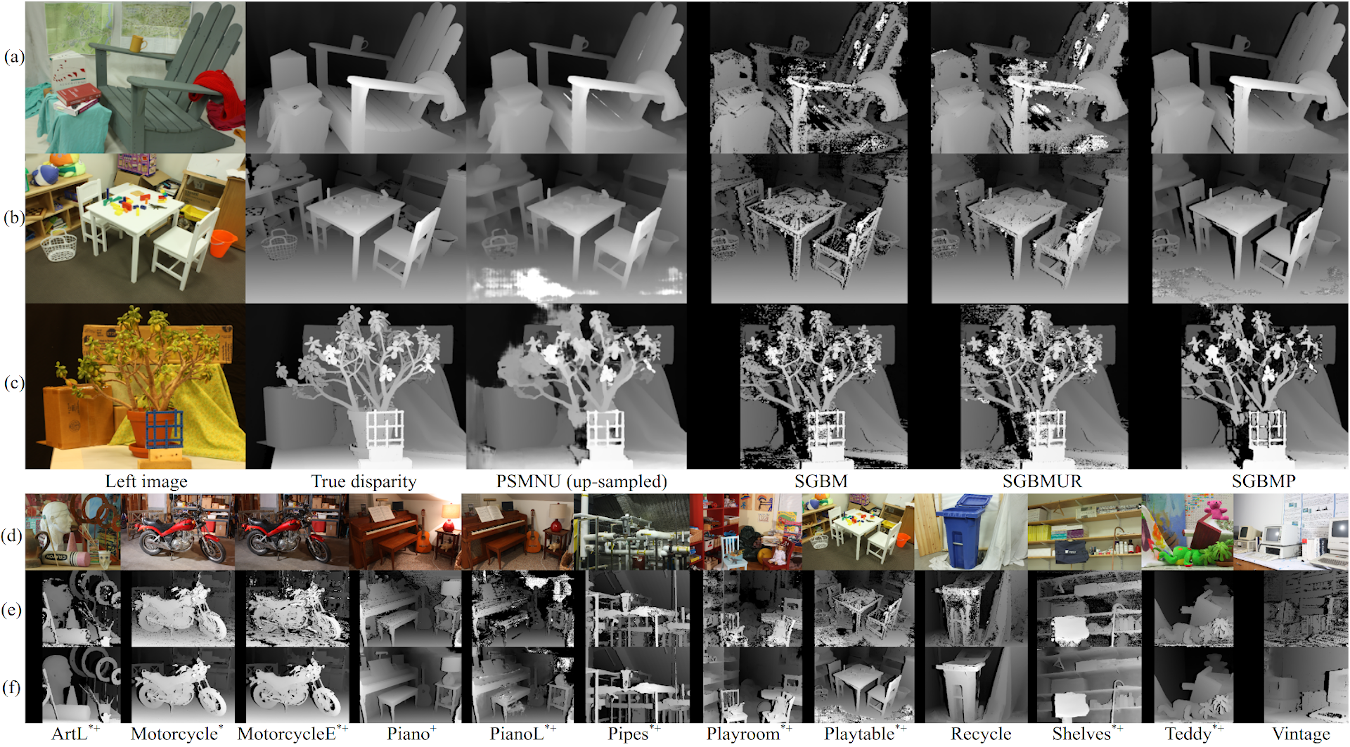}
    \caption{Comparison on the Middlebury dataset. Rows: (a) Adirondack case, (b) PlaytableP case, (c) Jadeplatn case, (d) left images, (e) disparities from SGBM, and (f) disparities from SGBMP. All of the disparity predictions are normalized with respect to the true disparity. In Row (a)-(c), the disparity prediction and $\sigma$ map from PSMNU is up-sampled to the original size of the input stereo images. Parameters for SGBM: \emph{SADWindowSize} 7,  \emph{P1} 1000,  \emph{P2} 4000,  \emph{disp12MaxDiff} 2,  \emph{preFilterCap} 31,  \emph{uniquenessRatio} 10 (0 for SGBMUR and SGBMP),  \emph{speckleWindowSize} 39,  \emph{speckleRange} 4. These parameters are chosen by experience without special fine tuning or grid search.}
    \label{fig:exp_comparison_row}
\end{figure*}

Among various openly available binocular stereo datasets, the Middlebury Stereo Evaluation V3 \cite{scharstein2014high} has large image size. We compared SGBM and SGBMP on all the 15 training cases of this dataset. The metrics of \emph{bad1.0}, \emph{invalid}, and \emph{avgErr} defined by Middlebury dataset are utilized. We also evaluate the standard deviation of \emph{avgErr}, denoted as \emph{stdErr}. \emph{stdErr} measures the noise level of a disparity prediction. Lower \emph{stdErr} means less noise. As discussed previously, we disable the UR check for SGBMP to make it possible to find a good stereo match inside difficult regions. In a second SGBM run, we also turn off the UR check for fair comparison and we name this run SGBMUR. The parameters are the same across the cases except the \emph{minDisparity} and \emph{numDisparity} \cite{OpenCV}. The parameters are listed in \reffig{fig:exp_comparison_row}. 

\reffig{fig:exp_comparison_row} shows results of all the tests while \reftab{tab:comparison_middlebury} lists the detailed values of the metrics associated to the (a)-(c) rows in \reffig{fig:exp_comparison_row}. 
SGBM invalidates many disparity predictions leading to high \emph{invalid} and low \emph{avgErr}, e.g. Jadeplant (\reffig{fig:exp_comparison_row} (c)). For our infrastructure inspection tasks, we prefer low \emph{invalid} and low \emph{avgErr}. PSMNU assists SGBMP to achieve this desired performance on 11 out of the 15 cases of the Middlebury dataset. Detailed results of all the 15 cases can be found on the project web-page\footnote{http://www.huyaoyu.com/technical/2019/09/09/Deep-assisted-high-resolution-binocular-stereo-reconstruction.html}. We select the results in \reffig{fig:exp_comparison_row} (a)-(c) and \reftab{tab:comparison_middlebury} to show three types of outcomes from SGBMP. 

\begin{itemize}
    \item \reffig{fig:exp_comparison_row} (a) Adirondack: When PSMNU performs well for most of the pixels, the prediction of SGBMP is better than or close to PSMNU.
    \item \reffig{fig:exp_comparison_row} (b) PlaytableP: If there are some regions where PSMNU makes large errors, SGBMP could compensate and give better disparity predictions. This could be illustrated in the floor region of the PlaytableP case. 
    \item \reffig{fig:exp_comparison_row} (c) Jadeplant: When PSMNU has a poor performance, SGBMP still manages to achieve similar \emph{avgErr} with SGBM but lower (and better) \emph{invalid} value. 
\end{itemize}


\begin{figure}[!b]
    \centering
    \includegraphics[width=3.2in]{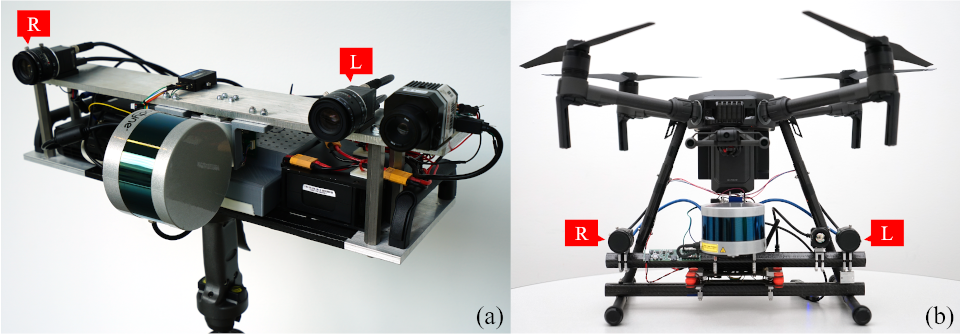}
    \caption{Customized stereo cameras. (a) Handheld platform. (b) UAV, payload has dampers for isolating vibrations. L, R are the 4K cameras  ($\text{3008}\times\text{4112}$). Images for this work are captured when the UAV is stationary.}
    \label{fig:exp_hardware}
\end{figure}

\begin{figure*}[!t]
    \centering
    \includegraphics[width=6.4in]{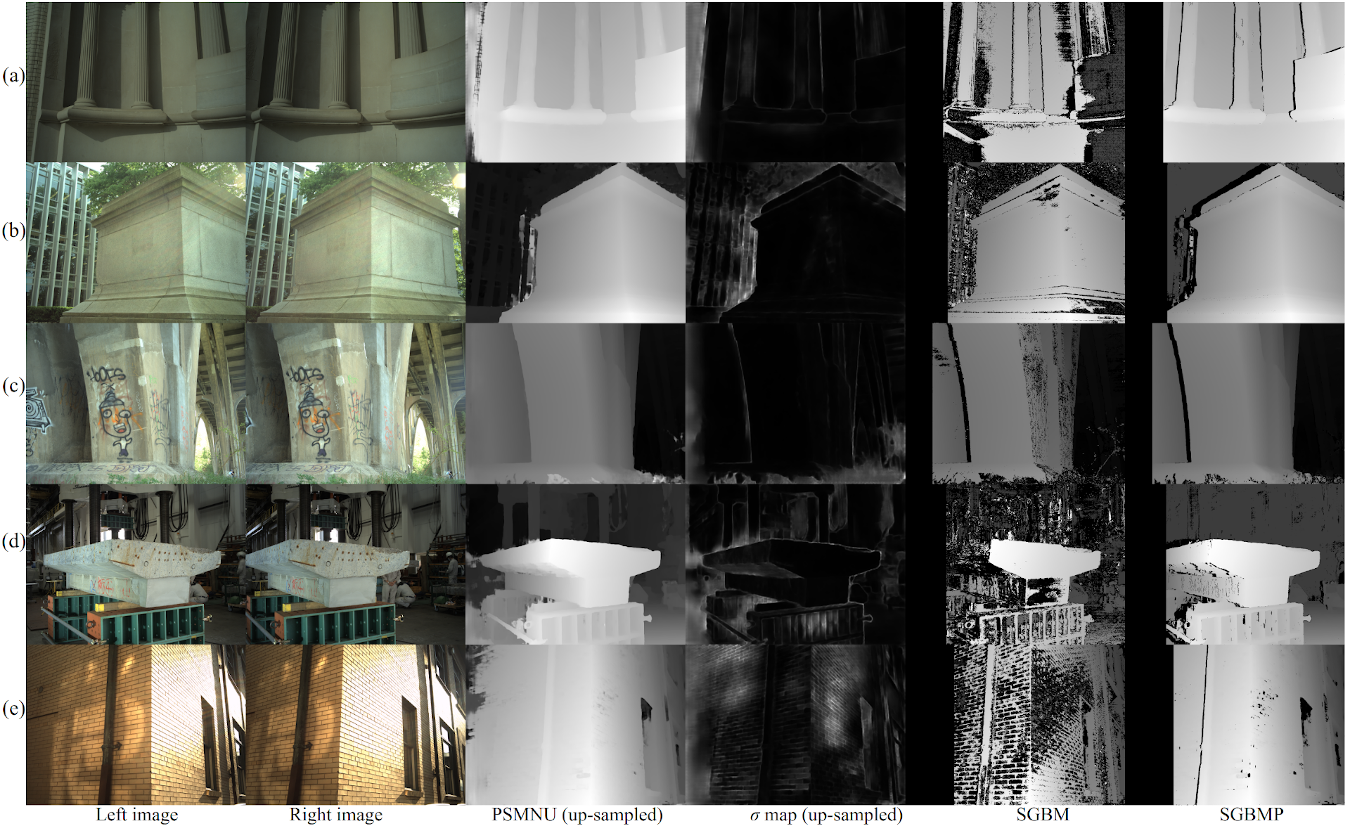}
    \caption{Comparison on the experimental data from infrastructure inspection tasks. Rows: (a) Stone gallery. (b) Concrete pillar. (c) Bridge support. (d) T-shaped beam. (e) Building walls. Execution time in seconds (SGBM/SGBMP): (a) 16.42 / 23.16, (b) 18.38 / 24.55, (c) 17.56 / 24.57, (d) 18.03 / 24.75, (e) 24.07 / 35.64. PSMNU's execution time is around 7s with an image size of $\text{752}\times\text{1028}$. Disparity predictions ($\mat{y}$) with uncertainties ($\sigma$ map) from PSMNU are up-sampled to the size of the original input stereo images. Disparity predictions inside a row are normalized. SGBM produces much fewer valid predictions for (a)(d)(e). Textures in (a) are uniform in general and the brightness and color are affected by the lens. The left-side surfaces of the T-shaped beam in (d) have an inconsistent color between the left and right images. Images in (e) have different levels of overexposure. Our SGBMP performs better on all these cases. }
    \label{fig:exp_comparison_realworld_5x6}
\end{figure*}

Results of all the other 12 training cases provided by Middlebury dataset are shown in Row (d)-(f) of \reffig{fig:exp_comparison_row}. Since PSMNU does not explicitly invalidate disparities, the \emph{invalid} values of PSMNU in \reftab{tab:comparison_middlebury} are the results of manually masking the left most disparity prediction and these values are for reference. The masked regions correspond to the areas in the left images where SGBM and SGBMP could not make any disparity predictions. Due to the nature of the difficult regions, such as textureless surfaces, disparity predictions may contain high level of noise despite having a low \emph{avgErr}. The \emph{stdErr} values in \reftab{tab:comparison_middlebury} evaluate the noise level. On 12 out of the 15 cases of the Middlebury dataset, SGBMP achieves the lowest \emph{stdErr} and \emph{invalid} at the same time (associated column names of Row (d)-(f) in \reffig{fig:exp_comparison_row} are marked by *). And 10 out of these 12 cases SGBMP have the lowest \emph{avgErr} (the column names are marked by +).

The down-sampled image size for PSMNU is $\text{768}\times\text{1024}$ for all the 15 cases. With this image size, PSMNU consumes around 8GB of GPU memory for a single pair of stereo images. \reftab{tab:comparison_middlebury} also shows the execution time of the Adirondack case as an example.  PSMNU's average execution time is about 1.5s. Based on our experimental results, SGBMP needs roughly 50\% more time than SGBM on average. We also submit all the results, including all the cases without ground truth, to the Middlebury Evaluation V3 web-site.

\subsection{Performance on real-world stereo images}

Tests on Middlebury dataset show promising performance gain of SGBMP. In this section, we test SGBMP on the high-resolution stereo images obtained by our experimental hardware in real-world infrastructure inspection scenarios. We use 4 identical 4K cameras to build 2 stereo cameras with identical baselines. These stereo cameras are installed to a handheld platform \cite{zhen2019joint} and a UAV as shown in \reffig{fig:exp_hardware}. However, for this work, images are captured without flying. The cameras are externally triggered and hardware-synchronized with other sensors such as the \LIDAR{} and IMU. We have to deal with many large and slanted surfaces and low texture regions. The lenses have a significant vignetting effect under low lighting conditions, making the left and right images have different brightness. We collected over 600 pairs of stereo images for 4 concrete structures and 2 building surfaces. In \reffig{fig:exp_comparison_realworld_5x6}, results from one camera position are shown for 5 test cases. The parameters adopted for SGBM and SGBMP are the same with \reffig{fig:exp_comparison_row} except 0 \emph{uniquenessRatio} for SGBMP and various \emph{minDisparity} and \emph{numDisparity}. \emph{minDisparity} and \emph{numDisparity} are selected individually for each test case to make sure that the true disparities are inside the selected ranges. The current computing resource forces PSMNU to work with the 1/16 of the original image size.

As illustrated in \reffig{fig:exp_comparison_realworld_5x6}, SGBMP improves the accuracy compared with SGBM, while achieving high-resolution results. This could be attributed to the robust performance of PSMNU on real-world data. Regarding the execution time, SGBMP also needs about 50\% more than SGBM.

\begin{figure*}[!t]
    \centering
    \includegraphics[width=6.4in]{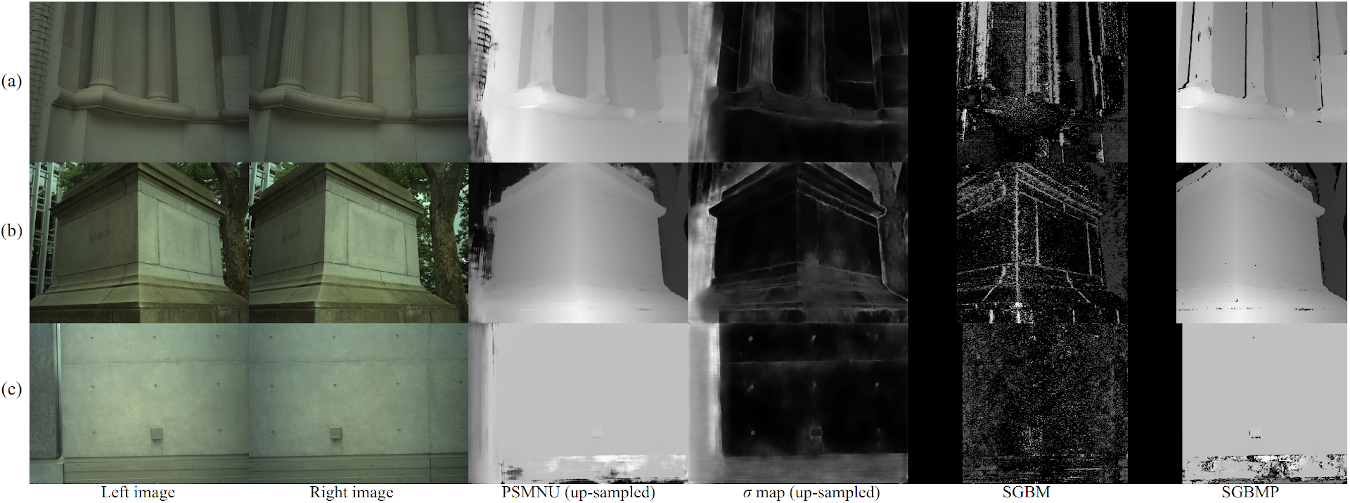}
    \caption{Comparison on the experimental data. Difficult cases for SGBM. Cameras were stationary. These stereo images originally have illumination and lens issues. Under insufficient illumination, the brightness of the images is low. Vignetting effects make the borders of the images even darker. Colors are inconsistent between the the two cameras, especially for (c). All the parameters used for these three cases are the same as \reffig{fig:exp_comparison_realworld_5x6}. SGBM only produces valid disparities at object boundaries in (a) and (b). Our SGBMP maintains its performance.}
    \label{fig:exp_comparison_realworld_3x6}
\end{figure*}

During the experiments, the camera may have a low performance with inadequate lighting. We show 3 such types of cases in \reffig{fig:exp_comparison_realworld_3x6} and they are extremely difficult for SGBM to do dense reconstruction. The Row (a) and (b) of \reffig{fig:exp_comparison_realworld_3x6} have roughly the same camera positions with Row (a) and (b) in \reffig{fig:exp_comparison_realworld_5x6}. However, the lighting conditions are worse and the images appear darker. In \reffig{fig:exp_comparison_realworld_3x6}~(c), the objective is simply a flat concrete wall with minor decorations. The brightness level is so low that we could observe the vignetting effects on the borders of each image. These stereo image pairs also have inconsistent color due to the lighting. Most of the valid disparity predictions of SGBM are around object boundaries. In contrast, PSMNU and SGBMP keep their performance and recover most of the pixels of the foreground objects with accurate disparities. 

\begin{figure}[htb]
    \centering
    \includegraphics[width=3.4in]{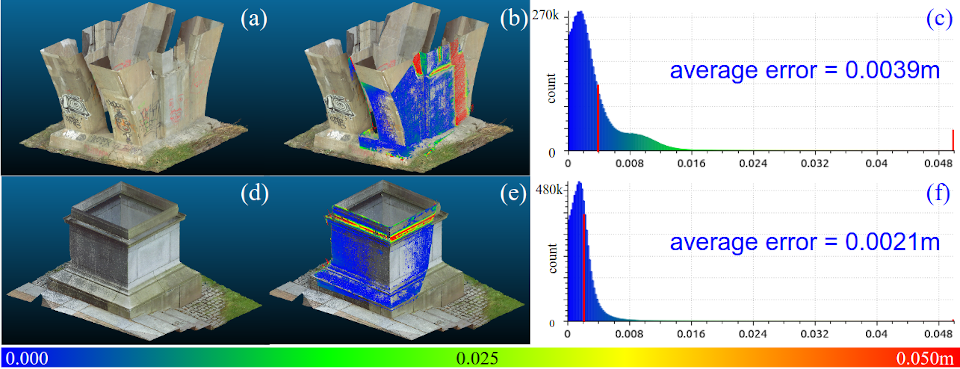}
    \caption{Point cloud comparison with the the survey scanner. (a) and (d): Point clouds from survey scanner. (b) and (e) Point cloud from SGBMP, colored by reconstruction error. (c) and (f): Error histograms. Reconstruction errors that are equal to or greater than 0.05m are rendered as red in (b) and (e). The long vertical red bars in (c) and (f) denote the corresponding average errors. SGBMP points that correspond to the pixels of the foreground object are shown and compared with the point cloud from the survey scanner.}
    \label{fig:exp_comparison_faro}
\end{figure}

Point clouds from a FARO $\text{FOCUS}^{\myss{3D}}$ survey scanner are utilized as the true depth \cite{zhen2019joint} to evaluate the absolute accuracy of SGBMP. We have scanned the stone pillar and the bridge support shown as the (b) and (c) rows in \reffig{fig:exp_comparison_realworld_5x6}. The camera poses are also obtained from \cite{zhen2019joint}. For every predicted 3D point in the SGBMP cloud, a plane is fitted by referring to the neighboring points from the survey scanner found within a radius of 0.05m. Then the point-to-plane distance is utilized as the reconstruction error. As shown in \reffig{fig:cover_figure} and \reffig{fig:exp_comparison_faro}, the average errors are lower than 0.004m with the majority of the predicted points having errors lower than 0.01m. We observe that the error of SGBMP becomes larger as the 3D points locate further away from the camera. Large errors occur near the object edges where large depth discontinuities and occlusions are present.

\section{Conclusions}
We present a high-resolution binocular stereo depth reconstruction pipeline by combining deep-learning model and a non-learning method.
Our deep-learning model, PSMNU, estimates its uncertainty on disparity prediction, and we use the uncertainty as a per-pixel searching range for the true disparity. With restricted computing resources, PSMNU produces accurate disparity prediction with associated uncertainties on down-sampled stereo images. The initial disparity prediction and the per-pixel disparity searching range are sent to the downstream non-learning method, SGBMP. SGBMP then predicts a dense disparity map with improved accuracy and high valid pixel rate on high-resolution stereo images. We evaluate our approach on the Middlebury Stereo Evaluation V3 dataset. SGBMP delivers superior accuracy over both SGBM and PSMNU for most of the cases. The absolute accuracy is also evaluated on our 4K infrastructure inspection images. We compare SGBMP with the point clouds collected by a survey scanner. The experiments show the average reconstruction error is below 0.004m.

Although we show significant improvements over various scenarios, the proposed method could still give bad predictions if PSMNU fails. To make PSMNU more robust, incorporating multi-task learning and multi-view depth reconstruction may be the way to explore in our future studies.






\section*{ACKNOWLEDGMENT}

Special thanks to Huai Yu (Wuhan University, China) and Daisuke Hayashi (Shimizu Corporation, Japan) for helping with the data collection. 


\bibliographystyle{./bibliography/IEEEtran}
\bibliography{./bibliography/bibliography,./bibliography/Supplementary,./bibliography/IEEEabrv,./bibliography/IEEEexample}

\end{document}